%% file: arxiv.tex
\documentclass{article}
\usepackage[T1]{fontenc}
\usepackage[english]{babel}
\usepackage{csquotes}

\newcommand{\ArXiv}[2]{#1}
\newcommand{\email}[1]{{\small\texttt{#1}}}
\newcommand{\inst}[1]{${}^{\textnormal{#1}}$}

\newcommand{\authorrunning}[1]{}
\newcommand{\institute}[1]{}
\newcommand{\orcidID}[1]{${}^{\textnormal{\href{https://orcid.org/#1}{[#1]}}}$}
\newcommand{\keywords}[1]{\\\;\\\textbf{Keywords: } \begingroup\newcommand{\and}{$\cdot$ }#1\endgroup}

\newenvironment{credits}{}{}
\newcommand{\ackname}{Acknowledgment.}
\newcommand{\discintname}{Disclosure of Interests.}

\usepackage[T1]{fontenc}
\usepackage{graphicx}
\usepackage{amsmath, amssymb, amsthm}

\theoremstyle{definition}
\newtheorem{definition}{Definition}

\theoremstyle{plain}
\newtheorem{lemma}{Lemma}

\newtheorem{theorem}{Theorem}

\theoremstyle{remark}

\renewenvironment{proof}[1]
  {\par\noindent\emph{Proof of #1.}\quad}
  {\hfill$\square$\par}

\newcommand{\toappendix}[1]{{#1}}
\usepackage{etoolbox}
\newcommand{\appendixbuffer}{}
\renewcommand{\toappendix}[1]{\appto\appendixbuffer{#1}}
\newcommand{\printappendix}{\appendixbuffer}

\usepackage{hyperref}
\usepackage[capitalize]{cleveref}
\usepackage{color}
\usepackage{tikz}
\usetikzlibrary{calc}
\usetikzlibrary{positioning}
\usepackage{algorithm}
\usepackage{algpseudocode}
\usepackage{subcaption}

\newcommand{\PT}{P_T}
\renewcommand{\d}{\textnormal{d}}
\newcommand{\F}{\mathcal{F}}
\newcommand{\Ft}{\overline{\F}}
\renewcommand{\t}{{f_\T}}
\newcommand{\R}{{\mathbb R}}
\newcommand{\X}{{\mathcal X}}
\newcommand{\D}{{\mathcal D}}
\newcommand{\T}{{\mathcal T}}

\newcommand{\Do}{\textnormal{do}}
\newcommand{\pa}{\textnormal{pa}}

\renewcommand{\P}{{\mathbb P}}
\newcommand{\indep}{\perp\!\!\!\perp}
\newcommand{\nindep}{{\not\indep}}

\newcommand{\emojiShower}{\includegraphics[height=\ht\strutbox]{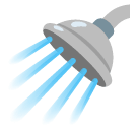}}
\newcommand{\emojiCloud}{\includegraphics[height=\ht\strutbox]{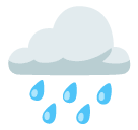}}
\newcommand{\emojiRain}{\includegraphics[height=\ht\strutbox]{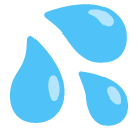}}

\begin{document}
\input{body}
\bibliographystyle{abbrv}
\bibliography{bib}
\appendix
\section{Proofs}
\printappendix
\end{document}

%% file: body.tex
\newcommand\blfootnote[1]{%
  \begingroup
  \renewcommand\thefootnote{}\footnote{#1}%
  \addtocounter{footnote}{-1}%
  \endgroup
}

\title{Causal Explanation of Concept Drift --\\ A Truly Actionable Approach\ArXiv{\:\:\footnote{This manuscript was presented at the TempXAI workshop at the  European Conference on Machine Learning and Principles and Practice of Knowledge Discovery in Databases (ECMLPKDD 2025).}}{}}
\author{
David Komnick\inst{1}${}^{,\dagger}$\and 
Kathrin Lammers\inst{1} \and 
Barbara Hammer\inst{1} \orcidID{0000-0002-0935-5591} \and 
Valerie Vaquet\inst{1} \orcidID{0000-0001-7659-857X} \and 
Fabian Hinder\inst{1} \orcidID{0000-0002-1199-4085}%
\ArXiv{ \\\;\\
Machine Learning Group, Bielefeld University, Bielefeld, Germany \\
\email{\{dkomnick,klammers,bhammer,vvaquet,fhinder\}@techfak.uni-bielefeld.de}
}{}
}
\institute{Machine Learning Group, Bielefeld University, Bielefeld, Germany \\
\email{\{dkomnick,klammers,bhammer,vvaquet,fhinder\}@techfak.uni-bielefeld.de}}
\authorrunning{D. Komnick et al.}
\maketitle   
\blfootnote{\!\!\!\!${}^\dagger$ Corresponding Author}          
\begin{abstract}

In a world that constantly changes, it is crucial to understand how those changes impact different systems, such as industrial manufacturing or critical infrastructure. Explaining critical changes, referred to as concept drift in the field of machine learning, is the first step towards enabling targeted interventions to avoid or correct model failures, as well as malfunctions and errors in the physical world. Therefore, in this work, we extend model-based drift explanations towards causal explanations, which increases the actionability of the provided explanations. We evaluate our explanation strategy on a number of use cases, demonstrating the practical usefulness of our framework, which isolates the causally relevant features impacted by concept drift and, thus, allows for targeted intervention.
\keywords{Concept Drift \and Explainable AI \and Computational Causality \and Model-based Drift Explanations \and Causal Explanations.}
\end{abstract}

\section{Introduction}

Machine learning plays a considerable role in many aspects of life, ranging from private usage through social media, chatbots, and recommender systems to many use cases in industry, e.g., quality control, etc. Despite many successful applications, many challenges arise when applying machine learning in critical applications. As identified in the European AI Act \cite{european_commission_and_directorate-general_for_communications_networks_and_content_and_technology_proposal_2021}, a key aspect is the so-called black box behavior of many machine learning models. With the increasing complexity of models, the rationale for their predictions has become increasingly opaque. To address this challenge, an entire field focusing on explainable AI (XAI) has emerged in the last decade \cite{adadi_peeking_2018,barredo_arrieta_explainable_2020,molnar2019}. The goal is to understand model decisions better and provide appropriate explanations to the users, potentially enabling suitable actions.

Beyond their original role of enhancing the interpretability of machine learning models, XAI methods increasingly demonstrate potential as tools for broader data analysis tasks. This can be particularly useful if the data is complex and time is scarce, as is the case in many real-world settings, where the underlying data distribution might change over time \cite{gama}. Here, a deeper understanding drift not only allows for the adaptation and improvement of stream learning algorithms, but is also crucial in \emph{system monitoring} where changes can indicate the necessity to take action, either by autonomous procedures or human operators overseeing complex systems.
Hence, next to the mere detection of drifts, a suitable explanation is frequently required \cite{partB}. In this paper, we will focus on the latter assignment. We propose to extend the framework by \cite{neucomp} to provide causal explanations that directly provide actionability.

This work is structured as follows. First, we recap the definition of concept drift (\cref{sec:drift_definition}) and summarize the related work on drift explanations (\cref{sec:relwork}). Before proposing causal drift explanations in \cref{sec:method}, we describe the required concepts from computational causality in \cref{sec:causality_definition}. We then evaluate the proposed explanation pipeline experimentally (\cref{sec:exp}) and conclude the paper (\cref{sec:concl}).

\section*{Notation}
In the following, we consider a data space $\X$ composed of multiple features. We refer to the index set of all features as $\F, | \F |  < \infty$. Every feature $f \in \F$ takes values in the real numbers $\R$, i.e., $\X = \R^\F$.
For a subset $F \subset \F$ we write $\X_F$ for the subspace based on the features in $F$, and for a data point $X$, we write $X_F$ for the projection onto $\X_F$. In addition, if $P$ is a probability measure on $\X$ we also write $P(X_F = x)$ or $P_{|\X_F}$ for the marginalization of $P$ onto $\X_F$.

\section{Concept Drift}
\label{sec:drift_definition}

In the classical batch learning setup, one assumes the data is given as random variables $X_1,\dots,X_n$ that are independent and identically distributed (iid) according to some probability distribution $\D$.
In contrast, in many real-world applications, we encounter the issue that the data is not identically distributed but subject to change -- a phenomenon referred to as concept drift~\cite{gama,lu}. This can be due to the course of time, as in stream learning~\cite{gama}; the data collection process taking place at different locations, as considered in the context of federated learning \cite{yang_federated_2020}; changes in the used equipment or sensors, such as sensor drift or applications of transfer learning \cite{10.5555/1462129}, or combinations thereof. 
As pointed out by \cite{partB}, formally, all these cases can be modelled using an abstract time domain, $\T$, that, for example, encodes clock time, the considered location, or computational node, and associating a -- potentially different -- distribution $\D_t$ to each abstract time point $t \in \T$. Concept drift refers to not all $\D_t$ being equivalent, i.e., there are $s,t \in \T$ such that $\D_t \neq \D_s$ \cite{gama,lu}.

In \cite{dawidd}, the authors suggested a statistical modelling explicitly including time. This allows for an equivalent formalization of drift as data and time being dependent, i.e., assuming the sample $X$ was observed at time point $T$ then we have $X \sim \D_T$ and there is drift if and only if $X$ and $T$ are not statistically independent. The advantage of this definition for algorithm development~\cite{neucomp2,waterleak,roberts2025conceptualizing} and in particular, drift explanations~\cite{neucomp,partB}, is that it encodes drift as a non-trivial relation of data and time~\cite{dawidd}.

\begin{definition}
    Let $\T$ be a time domain, and $\X = \R^d$ be a data space. We say that the distribution process~\cite{partB} $(\PT,\D_t)$, i.e., a probability measure $\PT$ on $\T$ together with a Markov kernel $\D_t$ from $\T$ to $\X$, has drift iff one of the following equivalent holds~\cite{dawidd}:
    \begin{enumerate}
        \item observing $\D_t \neq \D_s$ with probability larger 0, i.e., $\PT^2(\{(s,t) \::\: \D_t \neq \D_s\}) > 0$
        \item data and time are not independent, i.e., for $T \sim \PT$ and $X \mid T = t \sim \D_t$ we have $\P[X \in A, T \in W] \neq \P[X \in A] \P[T \in W]$
    \end{enumerate}
    We refer to the joint distribution of $X$ and $T$, i.e., $\P[X \in A, T \in W] = \int_W \D_t(A) \d \PT(t)$, as the \emph{holistic distribution}.
\end{definition}

In the next section, we discuss the related work on explaining drift and causality in the drifting setup.

\section{Related Work}\label{sec:relwork}
Understanding drift is of major importance in many scenarios as it enables operators to perform interventions in the system at hand or to adapt models. Still, research on this topic, especially with regard to actionable explanations, is still limited. Some works focus on detecting and quantifying the drift, while others attempt to visualize it \cite{partB}. Besides, some works provide feature-wise explanations of concept drift~\cite{webb2017understanding,neucomp,lu,partB}.

A particularly versatile framework for drift explanations is \emph{model-based drift explanations} \cite{neucomp}. This family relies on modelling drift as a relation of data and time as introduced in \cref{sec:drift_definition}. They employ learning models as surrogates to compute explanations describing the drift. In this framework, a suitable model is trained to predict the time point $T$ based on the sample $X$. Afterwards, the model is analysed using common, generic explanation methods ranging from interpretable models~\cite{neucomp}, over feature relevance~\cite{esann2023,neucomp2,esann2024,waterleak}, to counterfactual explanations~\cite{esann2022}, and activation vectors~\cite{roberts2025conceptualizing}. 

While the model-based explanation framework is very versatile as it works with generic explanation methods, so far, there has been a focus on exploratory explanation techniques~\cite{molnar2019}. These constitute a possibility to get an insight into how the data stream is changing overall. However, in many settings, human users require more actionable explanations. Since it is natural for many people to think about the cause and effect of an observation, some kind of causation-based explanations would be desirable \cite{cheng_covariation_1997}. 

Work on causal explanations of drift is very limited. A few publications focus on related tasks, e.g., forecasting \cite{chihara_modeling_2025} or drift detection \cite{yang_detecting_2025}, and only discuss explanations in passing. They propose finding two causal models based on directed acyclic graphs (DAGs), whereby one represents the data collected before the drift and a second one that collected after the drift. The causal explanation is derived from the difference between the causal models. While \cite{chihara_modeling_2025} extracts the causal structures from in an online fashion, \cite{yang_detecting_2025} relies on the so-called NOTEARS causal discovery algorithm. There are further contributions considering the intersection of drift and causality, e.g., \cite{baier_utilizing_2020,10.5555/3042573.3042635}. However, these are disjoint from our research question.

While some works consider the related field of feature relevance theory for explaining drift \cite{siirtola2023feature,neucomp2}, in this work, we aim to extend the model-based drift explanation framework to provide actionable causal explanations of drift. Before deriving our methodology (\cref{sec:method}), we will recall the most important aspects from computational causality.

\section{Computational Causality}
\label{sec:causality_definition}

Causality as a concept lies at the core of human reasoning. It shapes our perception, decision making, and how we predict outcomes \cite{cheng_covariation_1997}. Despite its central role in understanding the world, there is no obvious way to formalize or model the intuitive concept of causality, leading to several different formalizations. Here, we will use the concept of computational causality induced by \emph{interventional $\Do$-calculus}~\cite{pearl} due to its model-based nature and resulting similarity to the model-based drift explanation framework. 

The backbone of this approach is given by \emph{Bayes networks}, a kind of probabilistic graphical model. Building up on an acyclic directed graph (DAG) $G$ with the set of nodes corresponding to the set of features, i.e., $V(G) = \F$, and the edges indicating the connections between the variables. More precisely, the distribution of a feature can be computed solely based on the values of its parents, i.e., $\P[X] = \prod_{f\in\F}\P[X_f \mid X_{\pa(f)}]$. We will refer to the collection of conditional distributions as $P_f = \P[X_f \mid X_{\pa(f)}]$, making the whole model $(G,P_f)$. Thus, the network can be seen as a computational graph to generate samples from a distribution, which we denote as $P_G$. We will use this model as a link to explore computational causality in the framework of model-based drift explanations. 

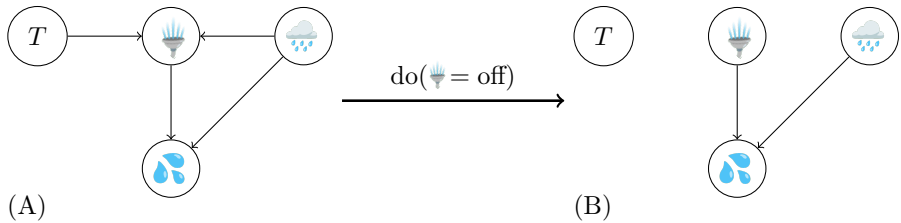
\begin{figure}[!t]
    \centering
    \resizebox{\textwidth}{!}{
    \begin{tikzpicture}
        \node (subfig1) {
            \begin{tikzpicture}
                \node[circle, draw] (A) {\:\quad\:};
                \node[circle, draw, right=of A] (B) {\:\quad\:};
                \node[circle, draw, below=of A] (C) {\:\quad\:};
                \node[circle, draw, left=of A] (T) {$\:T\:$};
                \node[right=-1.1em of A.center] {\!\rotatebox[origin=c]{225}{\Large\emojiShower}\!};
                \node[right=-1.1em of B.center] {\Large\emojiCloud};
                \node[right=-1.1em of C.center] {\Large\emojiRain};

                \draw[->] (A) -- (C);
                \draw[->] (B) -- (A);
                \draw[->] (B) -- (C);
                \draw[->] (T) -- (A);
            \end{tikzpicture}
        };
        \node[right=0cm of subfig1.south west] {{(A)}};

        \node[right=3cm of subfig1] (subfig2) {
            \begin{tikzpicture}
                \node[circle, draw] (A) {\:\quad\:};
                \node[circle, draw, right=of A] (B) {\:\quad\:};
                \node[circle, draw, below=of A] (C) {\:\quad\:};
                \node[circle, draw, left=of A] (T) {$\:T\:$};
                \node[right=-1.1em of A.center] {\!\rotatebox[origin=c]{225}{\Large\emojiShower}\!};
                \node[right=-1.1em of B.center] {\Large\emojiCloud};
                \node[right=-1.1em of C.center] {\Large\emojiRain};

                \draw[->] (A) -- (C);
                \draw[->] (B) -- (C);
            \end{tikzpicture}
        };
        \node[right=0cm of subfig2.south west] {{(B)}};
        \draw[line width=1pt, ->] (subfig1) to node[above] {$\Do(\text{\!\!\rotatebox[origin=c]{225}{\emojiShower}\!\!} = \text{off})$} (subfig2);
    \end{tikzpicture}}
    \caption{
    Simple causal model (A) with time ($T$) controlled sprinkler (\!\!\rotatebox[origin=c]{225}{\emojiShower}\!\!) which also switches off if it rains (\emojiCloud); in both cases the ground will be wet (\emojiRain). If we manually turn off the sprinkler (B), i.e., apply a $\Do$-operator, all connections to its parents are removed in the graphical model.
    }
    \label{fig:do}
\end{figure}

The connection of such models to the intuitive notion of causality is given by interpreting the directions of the edges as the direction from cause to effect. Yet, as every order of features induces a DAG by conditioning on all previous features, this is insufficient. This gap is bridged by the \emph{$\Do$-operator}: usually the cause-variable affects the effect-variable but not the other way around -- turning on the sprinkler will cause the road to be wet; making the road wet will not cause the sprinkler to be turned on (see \cref{fig:do} (A) for an illustration). This idea can be exploited to model a causal structure by linking it to experimental interventions: in an experiment, we force certain variables to take on specific values and observe the remaining ones, which can be formalized as follows:

\begin{definition}[Experiment with Interventions]\label{def:expIntervention}
Let $\X$ be a dataspace with features $\F$. An experiment with intervention $E$ is a (collection of) Markov kernels that takes a manipulation $x$ on features $F$ to an observed experimental outcome on the remaining features, i.e., for all $F \subseteq\F$ we have a map
\begin{align*}
    E_{F}: \X_F \times \Sigma_{\X_{\F \setminus F}} \to \R 
\end{align*}
such that $x \mapsto E_F(x,A)$ is measurable and $A \mapsto E_F(x,A)$ is a probability measure for all $x$ and $A$, respectively. We call $E_{F}(x)$ the \emph{$F$-manipulation of $E$ to $x$} and refer to $E_{\emptyset}(*)$ as the \emph{not-manipulated experiment}. (Here we set $\X_\emptyset = \{*\}$)
\end{definition}

In Bayes networks, we can simulate this idea by forcing a feature or a group of features into a specific state, $X_F = x$, and inferring the values of the remaining features. However, due to the asymmetry of cause and effect, we have to remove the paths leading to the manipulated features -- as those relate to causes -- while still keeping the paths from them -- as those relate to effects. We have illustrated this in the case of the sprinkler example in \cref{fig:do}. This kind of operation is called an intervention or the $\Do$-operation on the Bayes network. If the Bayes network coincides with reality for all such manipulations, it is called a \emph{causal model}.
This can be formalized as follows:
\begin{definition}
\label{def:causal_model}
Let $\X$ be a dataspace with features $\F$, $E$ be an experiment with intervention, and $(G,P_f)$ a Bayes network. We call $(G,P_f)$ a \emph{causal model (of $E$)} if for any intervention the experimental result and the model prediction coincide, i.e., for all $F \subset \F$ and $E_\emptyset(*)_{|\X_F}$-a.e. $x \in \X_F$ and $A \in \Sigma_{\X_{\F \setminus F}}$ it holds
\begin{align*}
    P_G( A \mid \Do(X_F = x)) = E_F(x,A)
\end{align*}
where $P_G(\cdot \mid  \Do(X_F = x)) := P_{G'}$ is the distribution of the model $(G',P')$ obtained by applying the $\Do$-operator, i.e., $(G',P_f') = \Do((G,P),F,x)$.
\end{definition}

It is of utmost importance not to confuse the $\Do$-operator with usual conditioning: while computing the distribution is equivalent to computing a conditional, this conditional is computed based on a network with the connection towards the manipulated variables removed (see \cref{fig:do} (B)). In other words, the $\Do$-operator coincides with usual conditioning only if applied to a set of variables that has no parents, i.e., we have
\begin{align*}
    P_G( \cdot \mid \Do(X_F = x)) = P_G(\cdot \mid X_{F} = x).
\end{align*}
Assuming that $P_G$ is the causal model for an experiment with intervention $E$, we can translate this idea back and define:

\begin{definition}[Causeless set of features]
    Let $E$ be an experiment with intervention with features $\F$.
    We say that $F \subset \F$ is \emph{causeless (in $E$)}, if 
    \begin{align*}
        E_F(x,A) = E_\emptyset(*,A \mid X_F = x),
    \end{align*}
    $E_\emptyset(*)_{|\X_F}$-a.s. for all $A$ where $E_\emptyset(*,A \mid X_F = x)$ denotes the conditioning of $E_\emptyset(*,\cdot)$ on $\X_F$\footnote{This is well-defined as $\X_F$ and $\X_{\F \setminus F}$ are standard Borel so that the conditional probability is regular and is uniquely determined.}.
\end{definition}

In this definition, the conditioning on the right-hand side can be considered as a way to process the data \emph{after} the experiment was performed. In other words, this definition entails the idea that we can train a non-causal model to predict the value of the remaining features from $x \in \X_F$ and still obtain an observation that fits the result of an intervention.

Algorithmically speaking, the objective of computational causality is to recover an underlying causal model from purely observational data. Concretely, given a dataset generated under the non‐intervention experiment $E_\emptyset(*)$, we want to infer the complete causal structure over all possible interventions in $E$. In practical terms, this approach aims to bypass the need to conduct every intervention experimentally. Achieving this reconstruction requires imposing assumptions that guarantee 1)~the existence of a valid causal model and 2)~computational tractability of the inference procedure. Typical assumptions are the Markov property, full information, or faithfulness that constrain the distribution of the observed variables and the considered setup, making it more similar to setups covered by those produced by Bayesian networks. A detailed discussion of these assumptions is beyond the scope of this paper. For our experimental evaluation, we focus exclusively on the classical PC algorithm~\cite{pearl}. 

\section{Causal Explanations of Concept Drift}
\label{sec:method}

In this section, we will derive the formal modelling of causal explanations for concept drift. To do so, we will proceed in three steps: 
1)~we link the formal statistical description of concept drift to the notion of experiments with intervention,
2)~formalize the task we want to solve in terms of an intervention, and then
3)~discuss how this can be realized in a practical application using the notion of computational causality.

\subsection{A Causal Model over Time}

Following the ideas of \cite{dawidd,neucomp}, that is, considering time as a feature with drift the dependency of time and data, allows us to easily model drift in the context of an experiment with intervention by simply extending the dataspace from $\X$ to $\X \times \T$, with the extended feature set $\Ft = \F \sqcup\{\t\}$, where $\t \not\in \F$ represents the specific time feature and we also write $X_{\t} = T$ for consistency.

To link drift to experiments with intervention, we assume that the passage of time is independent of our actions. We can formalize this by assuming that $\t$ is a causeless feature. 
Using this assumption, we can describe drift as follows:

\begin{theorem}
    \label{thm:link}
    Let $\X$ be a dataspace with features $\F$, $\T$ be a time domain, and $(\PT,\D_t)$ be a distribution process. Let $E$ be experiment with intervention on $\X \times \T$ that has the holistic distribution of $\D_t$ as the distribution of the not-manipulated experiment, i.e., $E_\emptyset(*,A \times W) = \int_W \D_t(A) \d \PT$, for which time ($\t$) is a causeless feature. Then the distributions $\D_t$ and the time-manipulations of $E$ coincide, i.e., 
    \begin{align*}
        \D_t = E_{\t}(t,\cdot)
    \end{align*}
    $\PT$-a.s. In particular, the presence of drift is equivalent to time-interventions affecting the data, i.e., 
    \begin{align*}
        \D_t \text{ has drift } \Longleftrightarrow E_{\t} \text{ is not constant ($\PT$-a.s.)}
    \end{align*}
\end{theorem}
\ArXiv{\toappendix{
    \begin{proof}{\cref{thm:link}}
        Since $\t$ is a causeless feature and $E_\emptyset(*,\cdot)$ is exactly the holistic distribution we have that the kernel $E_{\t}$ is the conditional of the holistic distribution on $\T$ which, by definition of the holistic distribution, is exactly $\D_t$ (as one choice). Then the fact that all considered spaces are standard Borel assures uniqueness. 
        
        That $\D_t$ has drift if and only if $\D_t = \D_\T$ $\PT$-a.s. was proven in~\cite{dawidd}.
    \end{proof}
}}{\begin{proof}All proofs are provided as part of the ArXiv version~\cite{arxiv}\end{proof}}
Note that we do not claim to perform an intervention on time here; we only state that if we were able to do this, we would observe the stated effect. 
Hence, it allows us to formalize the type of explanation we want to obtain in terms of actionability. This is a significant difference from most explanations for which no formal definition regarding their actual interpretation can be given~\cite{molnar2019}. More precisely, a \emph{causal drift explanation} is given by the manipulation we have to perform to reverse the drift:

\begin{definition}[Drift-Reversing Intervention]\label{def:dri}
    Let $\X$ be a dataspace with features $\F$, $\T$ be a time domain, and $(\PT,\D_t)$ be a distribution process with associated experiment with intervention $E$. Assume time ($\t$) is causless. 
    We say that $F\subseteq\F$ provides a \emph{drift-reversing intervention}, iff
    \begin{align*}
    \int_B E_F(x, A\times W) \d \D_t(X_F = x) = E_{\t}(t,A \times B) \PT(W)\quad \text{for $\PT$-a.e. $t$} 
\end{align*}
    for all $A \in \Sigma_{\X_{\F \setminus F}},\, B \in \Sigma_{\X_F},\,W\in \Sigma_\T$.
\end{definition}

In other words, a drift-reversing intervention requires that by controlling the values of the features in $F$ only -- which might be complicated but is not impossible -- we create the same effect as if we change the flow of time -- which is practically infeasible. Notice that here, we only need to specify the features we are about to alter, as the distribution is already forced upon us to match the time point-specific distribution. 

\subsection{First Analysis and Limitations} 
Using ideas from computational causality, we will derive a practical procedure for causal drift explanations.
To do so, in the following we will always assume that we consider an experiment with intervention $E$ that on the one hand is linked to a distribution process $(\D_t,\PT)$ via the holistic distribution as in \cref{thm:link,def:dri} and on the other hand that we have a causal model $(G,P_f)$ of $E$ (in the sense of \cref{def:causal_model}).

Using this setup, because $(G,P_f)$ is a causal model of $E$, performing the intervention on $E$ corresponds to computing the $\Do$-operator on $(G,P_f)$. Therefore, we can rephrase \cref{def:dri} in terms of the causal model. However, this time we will invoke the notion of time windows, which play a vital role in the analysis of stream learning algorithms:

\begin{lemma}
    \label{lem:dri_in_cm}
    Let $(\PT,\D_t)$ be a distribution process with a corresponding experiment with intervention $E$ on $\X \times \T$ with $\t$ a causeless feature. Let $(G,P_f)$ be a causal model of $E$. Then $F \subseteq \F$ is a drift-reversing intervention if and only if for all $A,B,W'$ and every time window $W \subseteq \T$, i.e., $\PT(W) > 0$, we have
    \begin{align*}
        \int_B P_G(X_R \in A, T \in W' \mid \Do(X_F = x)) \d P_G(X_F = x \mid T \in W) \qquad\qquad\qquad
        \\= \PT(W') P_G(X_R \in A, X_F \in B \mid \Do(T \in W)).
    \end{align*}
\end{lemma}
\ArXiv{\toappendix{
    \begin{proof}{\cref{lem:dri_in_cm}}
        By assumption 
        \begin{align*}
           P_G(A \times B \mid \Do(T \in W)) 
           &= \int P_G(A \times B \mid \Do(T = t)) \d P_G(T=t \mid T \in W) \\&= \int E_{\t}(t, A \times B) \d \PT(t \mid W) \qquad\qquad\qquad \text{ and }
        \end{align*}\vspace{-2em}\begin{align*}
           \int_B P_G(A \times W' \mid \Do(X_F = x)) \d &P_G(X_F = x \mid T
           \in W) 
           \\= \iint_B &E_F(x, A \times W') \d \D_t(X_F = x) \d \PT(t\mid W),
        \end{align*}
        using Fubini.
        Multiplying both sides by $\PT(W)$ the statement reduces to 
        \begin{align*}
           \int_W f \d P(t) = \int_W g \d P(t) \forall W \Leftrightarrow \PT[f = g] = 1
        \end{align*}
        which can be seen by subtracting the left-hand side and considering $W = \{f > g\}$.
    \end{proof}
}}{}

This again shows that the notion of drift-reversing interventions targets to reverse the drift; in this particular case, by ensuring that the distribution we currently observe is exactly the same as that observed during the time window $W$. The advantage of this formulation is that the result is more tangible, as window mean distributions play an important role in the context of concept drift~\cite{partB,neucomp2,dawidd,webb2017understanding,gama,lu}.

The great benefit of the additional structure of the causal model is that, by analysing its graphical structure, we can determine a drift-reversing set:

\begin{theorem}
    \label{thm:main}
    Let $(\PT,\D_t)$ be a distribution process with a corresponding experiment with intervention $E$ on $\X \times \T$ with $\t$ a causeless feature. Let $(G,P_f)$ be a faithful causal model of $E$. It holds
    \begin{enumerate}
        \item The node in $G$ corresponding to time, $\t$, has no parents
        \item The node time node has children if and only if $\D_t$ has drift
        \item Every drift-reversing set $F$ contains all children of $\t$
        \item The set of all children of $\t$ and their ancestors (without $\t$) are a drift-reversing set
    \end{enumerate}
\end{theorem}
\ArXiv{\toappendix{ 
     \begin{proof}{\cref{thm:main}}
         1. Follows directly from $\t$ being causeless. For any time window $W$ with positive measure and $A\in\Sigma_\F$, we have
         \[\int_W P(X\in S\mid T=x)\d P_T = P(X\in A\mid T\in W)\]
        and therefore $\t$ can be chosen as the first node in the graph node ordering of $G$, $\t$ has therefore no parents.
         
         2. In a faithful DAG a feature is a connected component in the graph if and only if the feature is independent of the rest\cite{pearl}. As $\pa(\t) = \emptyset$ we have $X \indep T$ if and only if there is some $f \in \F$ such that $\t \in \pa(f)$.

        3. Let $F$ a drift reversing set and assume there is a $f \not\in F$ with $\t \in \pa(f)$. Consider $A = \{x \in \X_R \::\: x_f \in A_f\}, W' = \T$ and multiply both sides of \cref{lem:dri_in_cm} by $P_G(T \in W)$.
        Then the right-hand side of \cref{lem:dri_in_cm} becomes 
        \begin{align*}
            &P_G(T \in \T)P_G(X_R \in A, X_F \in B \mid \Do(T \in W))P_G(W)
            \\&= P_G(X_f \in A_f, X_F \in B \mid \Do(T \in W))P_G(W)
            \\&= P_G(X_f \in A_f, X_F \in B \mid T \in W)P_G(W)
            \\&= P_G(X_f \in A_f, X_F \in B, T \in W)
            \\&= \int_{B \times W} P_G(X_f \in A_f \mid X_F = x, T = t) \d P_G(X_F = x, T = t),
        \end{align*}
        where the first equality follows because $T$ has no parents in $G$.
        Denote by $G'$ the graph obtained by $\Do(X_F)$, then left-hand side becomes
        \begin{align*}
        &\int_{B} P_G(X_R \in A, T \in \T \mid \Do(X_F = x)) \d P_G(X_F = x \mid T \in W) P_G(T \in W)
        \\&=\int_{B \times W} P_G(X_f \in A_f \mid \Do(X_F = x)) \d P_G(X_F = x, T = t)
        \\&=\int_{B \times W} P_{G'}(X_f \in A_f \mid X_F = x) \d P_G(X_F = x, T = t)
        \end{align*}
        Let $\mu$ be the bounded signed measure defined by
        \begin{align*}
            \mu(C) = &\int_{C} P_G(X_f \in A_f \mid X_F = x, T = t) \\&\qquad\qquad-  P_{G'}(X_f \in A_f \mid X_F = x)\d P_G(X_F = x, T = t).
        \end{align*}
        As $\mu$ is obtained by subtracting both sides, we have $\mu(B \times W) = 0$ and since those form an intersection stable generator of the $\sigma$-algebra we have $\mu = 0$ by the usual $\pi$-$\lambda$-argument.
        Hence $P_{G'}(X_f \in A_f \mid X_F = x) = P_G(X_f \in A_f \mid X_F = x, T = t)$ $P_G(X_F,T)$-a.s. But $P_{G'}(X_f \in A_f \mid X_F = x)$ is $t$-invariant so $P_G(X_f \in A_f \mid X_F = x, T = t) = P_G(X_f \in A_f \mid X_F = x)$. Therefore, 
        \begin{align*}
            & \int_B P_G(X_f \in A_f, T \in W \mid X_F = x) \d P_G(X_F = x)
            \\&= \int_B \int_W P_G(X_f \in A_f \mid T = t, X_F = x) \d P_G(T = t \mid X_F = x) \d P_G(X_F = x)
            \\&= \int_B \int_W P_G(X_f \in A_f \mid X_F = x) \d P_G(T = t \mid X_F = x) \d P_G(X_F = x)
            \\&= \int_B P_G(X_f \in A_f \mid X_F = x) P_G(T \in W \mid X_F = x) \d P_G(X_F = x)
        \end{align*}
        for all $B$ and thus
        \begin{align*}
            P_G(X_f \in A_f \mid X_F = x) P_G(T \in W \mid X_F = x) = P_G(X_f \in A_f, T \in W \mid X_F = x)
        \end{align*}
        so $X_f \indep T \mid X_F$ in $P_G$. Hence $X_F$ $d$-separates $X_f$ and $T$ which is a contradiction to $G$ being faithful.
        
        4. The set of all children of $\t$ and together with their parents form the Markov boundary of $\t$ so it $d$-separates $\t$ from the remaining graph. Hence, conditioning on this set allows us to set the system state. On the other hand, a set of features $F$ that is closed under taking parents, i.e., $\pa(f) \subset F$ for all $f\in F$, is causeless and thus $\Do$-operations and conditioning coincide. 
     \end{proof}
}}{}

In the next section, we will examine this result more closely and discuss a more refined notion of drift-reversing sets. 

\subsection{A Refined Approach}

\cref{thm:main} shows that the set of all children of the time node, together with all of their ancestors, forms a drift-reversing set. At first glance, this result produces a too large drift-reversing set as it is not to be expected that a far ancestor of a child of the time node needs to be contained in the drift-reversing set. 

To make this more explicit, consider the sprinkler example from the beginning visualized in \cref{fig:do}. As time ($T$) is already involved, we can apply our framework directly. We want to know why the street is dry in the evening: it is not raining, and the sprinkler is off, with the sprinkler being off because it is late. 
This suggests that the action we should invoke is to turn the sprinkler on, i.e., the sprinkler is the only drift-reverting feature. 

In contrast, according to \cref{thm:main}, we must also include the weather. On closer inspection (see \cref{fig:do} (B)), this is reasonable, as always turning on the sprinkler independent of whether or not it is raining causes another drift in the correlation of sprinkler and weather. Hence, to avoid this, we have to include the weather as stated by the theorem. 

We can get a more natural explanation, i.e., only the sprinkler, if we allow the intervention to depend on other values (the weather).  
The idea is that there is a causal core set of features given by all of the time node's children, and the intervention on those is allowed to depend on the value of their parents. This can be made explicit by asking which node distribution needs to be changed to ensure the global distribution is time-reversed:
\begin{theorem}
    \label{thm:minimal}
    In the setup of \cref{thm:main}, the smallest set of features that needs to be altered to ensure that we can obtain every time window distribution is exactly given by the set of all children of $\t$. In other words: for every window $W \subset \T,\, \PT(W) > 0$ there is a graphical model $(G_W,P_{W,f})$ on $\X$ such that $P_G(X \in A \mid \Do(T \in W)) = P_{G_W}(A)$ and $P_{W,f} = P_f$ for all $\t \not\in \pa_G(f)$. Conversely, for every $f \in \F$ that has $\t$ as a parent, there exist two time windows $W,W'$ such that $P_{W,f}\neq P_{W',f}$. 
\end{theorem}
\ArXiv{\toappendix{
    \begin{proof}{\cref{thm:minimal}}
        Without loss of generality we can assume that $A=A_{f_1}\times\ldots\times A_{f_n}$ with $P_G(X \in A) 
        > 0$. Define $C$ as the set of all children of $\t$ in $G$ and $O := \F\setminus C$, the other features.
        Fix $W\in \Sigma_\t$ with $P(T\in W, X \in A) > 0$.

        Choose any topological order $\sigma$ of $G$.  Create $G'$ by
        (i) adding $g\!\to\!f$ for all distinct $g,f\in C$ with
        $\sigma(g)<\sigma(f)$ and
               $f\in C$,\\[6pt]
          $P_G\!\bigl(x_f\mid x_{\pa_{G'}(f)}\bigr)$,
        (ii) deleting the time node $\t$ and all its incident edges.
        
        For each feature $f\in\F$, define
        \begin{align*}
        q_f\!\bigl(x_f\mid x_{\pa_{G'}(f)}\bigr)
        :=
          P_G\!\bigl(x_f\mid x_{\pa_{G'}(f)},\,T\in W\bigr)
               &\qquad \textnormal{ for } f\in O,
        \end{align*}
        where $I_f=\{g\in C:\sigma(g)<\sigma(f)\}$ and
        $\pa_{G'}(f)= (\pa_G(f)\setminus\{\t\})\cup I_f$ for $f\in C$.
        The Kernels provide a new distribution $Q$.

        Since \(G'|_{C}\) is fully connected, every earlier child is a parent of
        each later one, so the product of the kernels \(q_f\;(f\in C)\) equals the
        chain–rule expansion of the conditional joint; hence
        \[
           Q(X_C\in A_C)=P_G(X_C\in A_C\mid T\in W).
        \]

        Observe that \(\pa_{G}(f)\setminus\{\t\}\subseteq\pa_{G'}(f)\) for every \(f\in C\); hence each kernel \(q_f(x_f\mid x_{\pa_{G'}(f)})\) conditions on \emph{all original parents of \(f\)} in addition to the new ones, so the information available for predicting \(X_f\) is only enlarged, never reduced, and no original parent–child dependency is lost.

        Since each kernel \(q_f\) is indexed by \(\pa_{G'}(f)\) and their product equals \(P_G(x\mid T\in W)\), the pair \((G',Q)\) is a graphical model of the conditional distribution \(P_G(X\in\,\cdot\mid T\in W)\).

        Regarding the minimality we fix $f\in \textnormal{ch}(\t)$. Let $U\subseteq\F\setminus\{f\}$ be the set of parents of $f$ in $G_W$, allowing any, but fixed, parents. Since $f$ is connected to $\t$ in $G$, a faithful graph, we have $X_f\nindep T\mid X_U$. We want to show, that there are $W,W'\in\Sigma_\t, A_f\in \Sigma_f$ and $A_U\in \Sigma_U$, such that 
        \[0\neq \int_{A_U} (P_{W,f}(A_f|a_U)-P_{W',f}(A_f\mid a_U)) \d P_{X_U}(a_U)\]
        holds. Hence $X_f\nindep T\mid X_U$ there exist $A,W$, such that
        \begin{align*}
            V(a_U) &:= P(X_f\in A_f\mid X_U = a_u)\\
            U(a_U) &:= P(T\in W\mid X_U = a_U)\\
            W(a_U) &:= P(X_f\in A_f, T\in W\mid X_U = a_U)
        \end{align*}
        such that the delta $D(a_u) := W(a_u)-V(a_u)U(a_u)$ is not 0 $P_U$-a.s., i.e. $0 \not\equiv D$.
        W.l.o.g., we assume that $S^+ := \{ a_U\in\X_{U} \mid D(a_U) > 0\}$ has a positive measure. Note, that on $S^+$ in addition $0 < V(a_U) < 1$ holds almost surely.

        By definition, we have
        \begin{align*}
            P_{W,f}(X_f\in A_f, X_U = a_U) &= \frac{P(X_f\in A_f\mid T\in W, X_U = a_U)}{P(X_f \in A_f\mid X_U = a_u)}=U(a_u) + \frac{D(a_u)}{V(a_u)}\\
            P_{W^C,f}(A_f, a_U) &= \frac{P(X_f\in A_f\mid T\in W^C, X_U = a_U)}{P(X_f \in A_f)} = U(a_U) - \frac{D(a_U)}{1 - V(a_U)}
        \end{align*}
        (which is well defined on $S^+$) and therefore, the difference of the kernels
        \[P_{W,f}(A_f, a_U) - P_{W^C,f}(A_f, a_U) = \frac{D(a_U)}{V(a_U)(1- V(a_U))}\]
        is positive on $S^+$, the kernels are not a.s. equal.
        
    \end{proof}
}}

In other words, if we keep track of the changes induced by the other parents, we only need to alter the direct children of the time node to reverse the drift.
Therefore, we get two kinds of explanations: the full intervention (as stated by \cref{thm:main}) consisting of the children of $T$ and all their ancestors, and the conditional intervention of the children only, which then has to take the other parents into account (as in \cref{thm:minimal}). 

This finding is very much in line with other findings from the literature. In~\cite{esann2023,neucomp2}, the authors considered model-based drift explanation~\cite{neucomp} through the lens of feature importance, showing that the resulting features can be seen as a wrapper method for feature selection for drift detectors. Later on, \cite{esann2024} extended on these ideas by introducing the notion of drift-inducing and faithfully drifting features, with the former identifying as those that ``induce'' the drift into the system and the latter ``following along''. The authors showed that in both cases, the found drifting features relate to relevant features\cite{john1994irrelevant} when time $T$ is considered as the target of conditional density estimation. This links drifting features closely to graphical models: following the ideas of \cite{polytime} the set of all drifting (or relevant) features corresponds exactly to the connected component of the skeleton containing the time node, while the drift inducing features (or strong relevant) relate to the Markov boundary of $T$, i.e., its children and their other parents -- similar to \cref{thm:minimal}. 

This consideration can be seen in two directions: on the one hand, it shows that even simple feature selection methods provide nearly causal explanations supporting their widespread usage in different applications~\cite{webb2017understanding,waterleak,neucomp2}, on the other hand, it connects our theoretical considerations to existing literature and methods that already have been successfully applied in a wide spread of real world applications, e.g., critical infrastructure like electrical grids~\cite{waterleak,webb2017understanding} and water distribution networks~\cite{neucomp,waterleak}, as well as land cover analysis~\cite{webb2017understanding}.

\begin{algorithm}[t]
\caption{Causal Explanation of Drift}\label{alg:algo}
\begin{algorithmic}[1]
\Function{ExplainDrift}{$S=((X_1,T_1),\dots)$ data stream}
    \State $G \gets \textsc{DetermineDAG}(S)$ \Comment{Run causal discovery algorithm, e.g., PC}
    \State $C \gets \textsc{GetChildren}(G,\t)$
    \State $P \gets \cup_{f \in C} \textsc{GetParents}(G,f) \setminus (\{\t\} \cup C)$
    \State $A \gets \cup_{f \in C} \textsc{GetAncesters}(G,f) \setminus \{\t\}$
    \State \Return $(C,P,A)$
\EndFunction
\end{algorithmic}
\end{algorithm}

The algorithmic solution for both cases, i.e., full drift-reversing intervention and conditional one, is presented in \cref{alg:algo}. In other words, the full intervention explanation (\cref{def:dri,thm:main}) is given by $A$, the conditional intervention (\cref{thm:minimal}) offered by the core set $C$ with the conditional on $P$. 

As can be seen, the algorithm mainly performs a causal discovery on the timed data points. The later steps only extract features according to their position in the graph. Therefore, the causal discovery dominates the runtime and memory complexity of the approach. Notice that this cannot be significantly reduced if we want to compute the full drift-reversing intervention, as in this case, we might have to explore the entire graph. Furthermore, while we will make use of the classical PC algorithm in our experiments, using any other causal discovery algorithm is also valid. 

\section{Empirical Evaluation}
\label{sec:exp}

In the following, we will empirically evaluate our considerations.\footnote{See \url{https://github.com/FabianHinder/DRAGON} for code and datasets.} First, we briefly summarize the datasets used and describe the experimental setup. Before presenting and analyzing the results for the drift explanations in \cref{sec:expdrift}, we perform a preliminary stability analysis of the causal discovery of the PC algorithm on the selected causal graphs (\cref{sec:respc}).

\subsection{Datasets and experimental setup}

For the empirical evaluation, we use semi-synthetic datasets sampled from Bayes nets, which were modified to create plausible drift scenarios~\cite{fairness_data_paper}. Based on the popular \emph{Adult} and \emph{Portuguese Student Performance} datasets, the inherent causal structure \cite{fairness_dataset_survey} was used to learn conditional probability distributions, which were modified for the following scenarios:

\begin{itemize}
    \item \emph{Adult Inflation}: inflation causes increases likelihood of high monetary values
    \item \emph{Adult Women in STEM}: women are more likely to work in STEM jobs; less likely to work in administrative fields
support    \item \emph{Student Girls Support}: female students are enrolled in support program
    \item \emph{Student Boys Support}: male students are enrolled in support program
\end{itemize}

Drifting data streams with abrupt concept drift were created by merging data sampled from the unmodified distributions before the drift with data sampled from a specific scenario distribution after the drift. For consistency, the drift point was set at 25.000 samples for all streams based on \emph{Adult}, with a total length of approximately 48.800 samples (differing slightly due to filtered-out missing values). For the \emph{Student Performance} based streams, the drift point was set after 2.000 samples with a total length of 5.000 samples per stream.

For each of the described scenarios, we perform 10 experimental runs. We first sample a stream and then evaluate the proposed methodology. For the PC algorithm, we use the default implementation from the \emph{causal-learn} Python package \cite{zheng2024causal} with the g-square test. We report and analyze the results in the remainder of this section. 

\subsection{Preliminary stability analysis\label{sec:respc}}

\begin{figure}[t]
    \begin{subfigure}[b]{0.5\textwidth}
        \includegraphics[width=\textwidth]{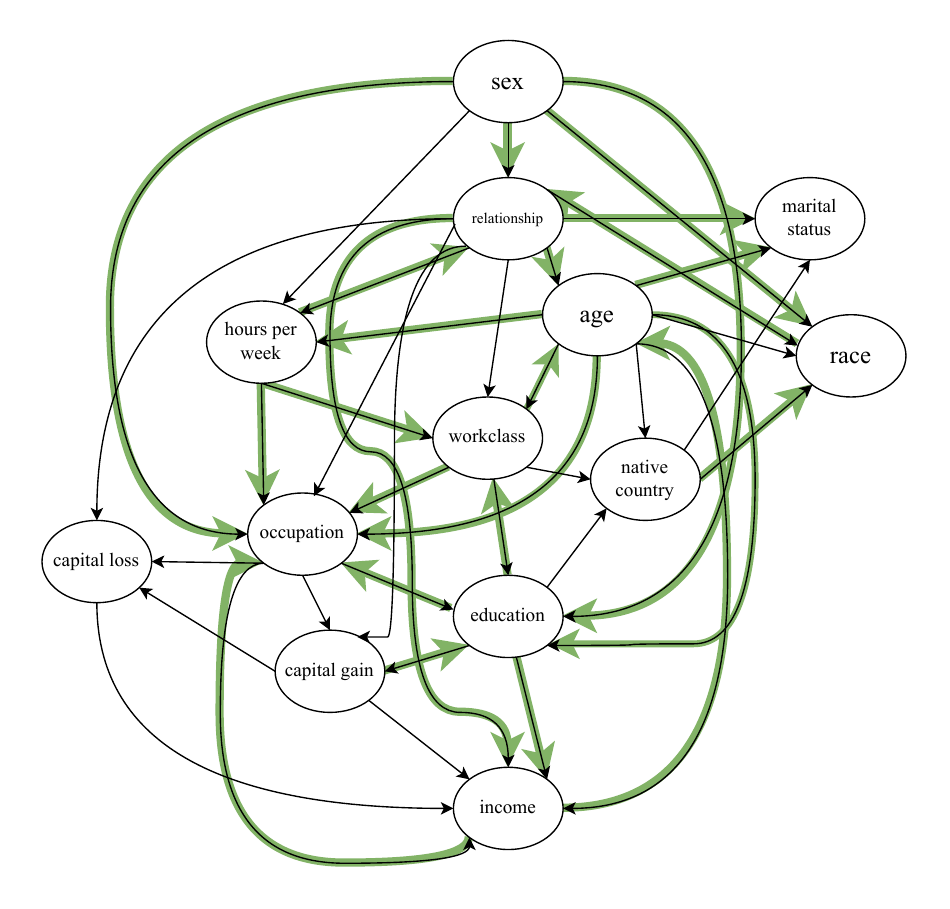}
        \hfill
        \caption{Adult}\label{fig:pc-adult}
    \end{subfigure}
    \begin{subfigure}[b]{0.5\textwidth}
        \includegraphics[width=\textwidth]{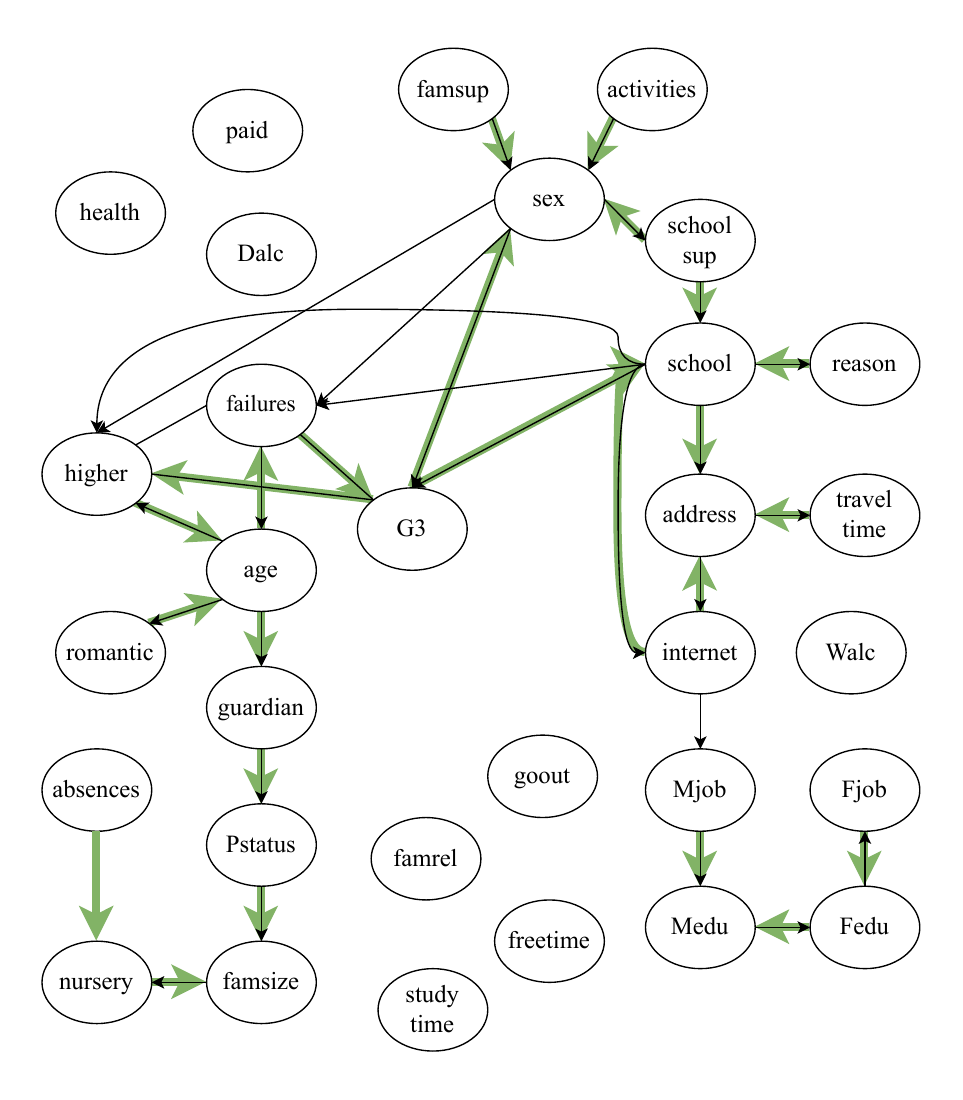}
        \caption{Student}\label{fig:pc-student}
    \end{subfigure}
    \caption{Performance of the PC Algorithm. Black edges indicate the ground truth, while green edges indicate detections by the PC algorithm; thickness correlates with the number of runs in which an edge was detected.}
    \label{fig:pc}
\end{figure}

In order to evaluate whether our causal explanations of drift are reasonable, we first evaluate the accuracy of the PC algorithm on the unmodified datasets -- those without temporal features.

Comparing the detected edges to the ground truth causal graph \cite{fairness_dataset_survey}, we find that the PC algorithm is moderately successful in detecting the causal structure underlying the unmodified \emph{Adult} dataset (\cref{fig:pc-adult}), where out of 38 edges in the ground truth, 19 (50\%) are detected correctly, while nine edges are wrongly oriented and ten were not detected at all. For the \emph{Student Performance} dataset (\cref{fig:pc-student}), meanwhile, the PC algorithm produces less accurate results, as it only detects eight out of 26 edges correctly (30.77\%). Additionally, 16 edges were detected but oriented inversely, while two edges were not detected, and three additional false edges were inserted.

The poor performance on \emph{Student Performance} can be explained by the comparatively low number of samples in this dataset, combined with a relatively high number of features. While the \emph{Adult} dataset only contains 13 features, whose connections can be learned from nearly fifty thousand samples, the \emph{Student Performance} dataset has 31 features and only 5.000 samples, which means that the PC algorithm has insufficient data to work with despite the lower connectivity in the causal graph, leading to less reliable independence tests.

\subsection{Experimental results\label{sec:expdrift}}
\begin{figure}[t]
    \begin{subfigure}[b]{0.5\textwidth}
        \includegraphics[width=\textwidth]{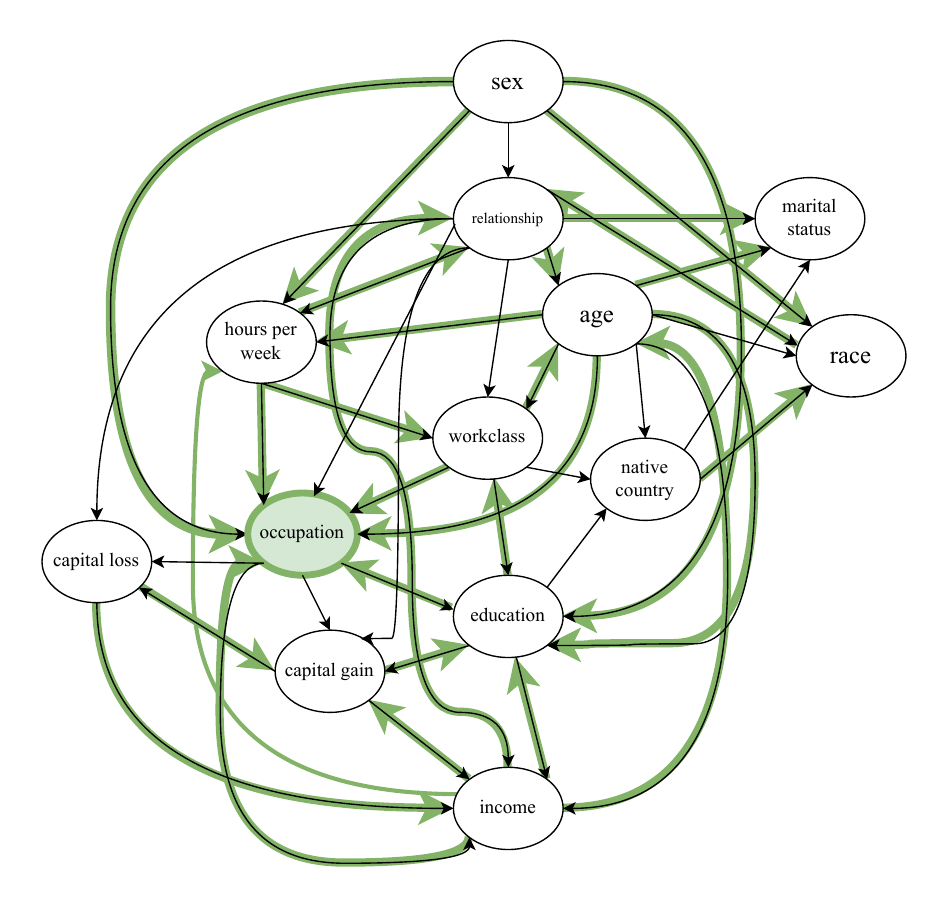}
        \caption{Adult -- Women in STEM}\label{fig:res-stem}
    \end{subfigure}
    \begin{subfigure}[b]{0.5\textwidth}
        \includegraphics[width=\textwidth]{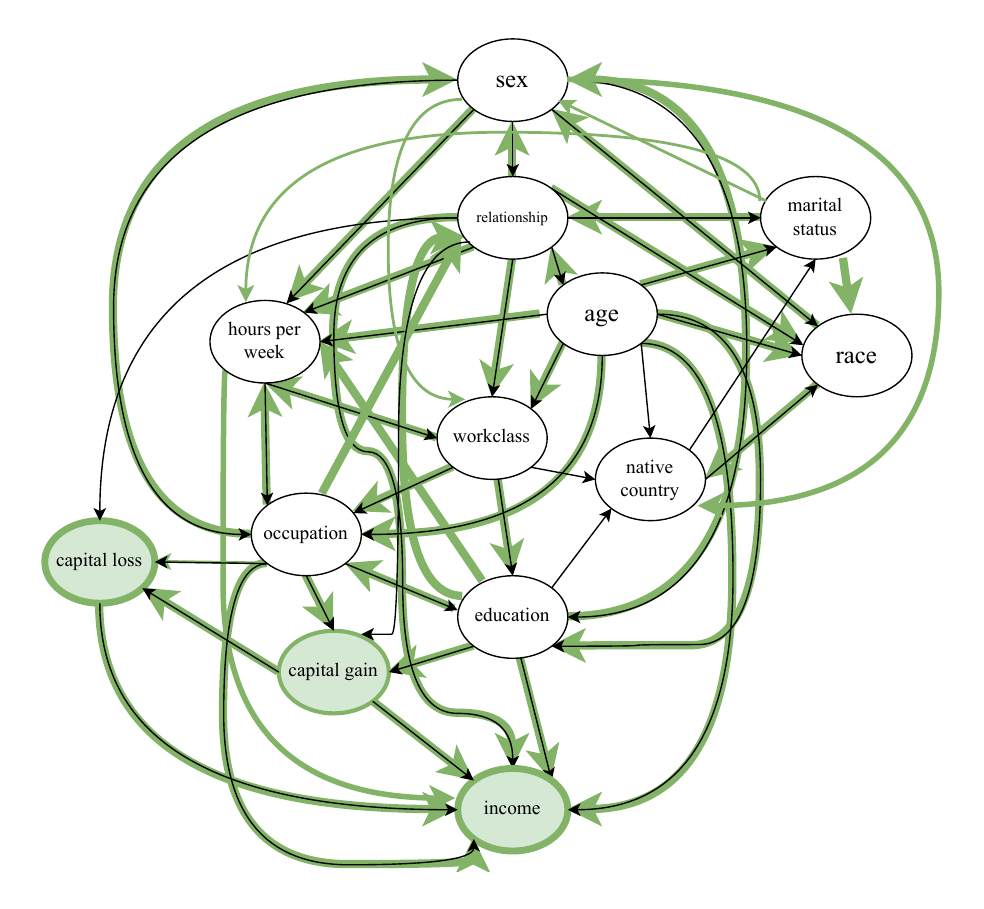}
        \caption{Adult -- Inflation}\label{fig:res-inflation}
    \end{subfigure}
    \caption{Case Studies on the Adult dataset. Black edges indicate the ground truth, while green edges indicate detections by the PC algorithm; thickness correlates with the number of runs in which an edge was detected. Children of $T$ are marked green, with the thickness of the border indicating the number of runs that correctly identified this relationship.}
\end{figure}
When analysing the causal structure of our drift scenarios, we can clearly see that our framework for causal explanations works -- the temporal feature $T$ is never connected with unrelated features, meaning that no relevant false explanations of drift are produced. And in the vast majority of cases, $T$ is correctly connected to the drifting feature, usually as a parent.

For our causal explanation, the feature(s) directly connected to $T$ are the most relevant, while other ancestors of children of $T$ -- i.e. possible conditional influences on the drifting feature(s) -- are also part of the explanation, though with less direct impact. These features are usually identified just as well in the drifting streams as on the unmodified data, which implies that potential limitations here lie with the PC algorithm as a causal discovery method, rather than with our theoretical framework.

To illustrate this, we first take a look at the \emph{Adult Women in STEM} drift scenario (\cref{fig:res-stem}), where the drifting feature gets reliably identified as \emph{occupation}. The temporal feature $T$ is detected as a parent of \emph{occupation} in nine out of ten experimental runs, while the other ancestors of the drifting feature are unchanged from the detected causal structure of the unmodified dataset. While we can see that our drifting stream results in overall more wrongly oriented edges in the causal graph, our drift explanation is generally stable and reliable. With the given information, a human data scientist would know just which feature's distribution to analyse further in order to fully understand the concept drift.

As the \emph{Adult Inflation} scenario (\cref{fig:res-inflation}) shows, these statements even hold true when there are multiple drifting features present in the stream. In eight out of ten experimental runs, all three drifting features -- \emph{capital-gain}, \emph{capital-loss}, and \emph{income} -- are correctly identified as directly connected to $T$. In the other two runs, only \emph{capital-gain} is not detected as a direct relation of the time, but this issue can be easily explained by the close relationship between the three drifting features, which leads to some issues with the independence tests employed by the PC algorithm. This close connection between the drifting features is also why the PC algorithm shows trouble orienting the temporal edge. 
\begin{figure}[t]
    \begin{subfigure}[b]{0.5\textwidth}
        \includegraphics[width=\textwidth]{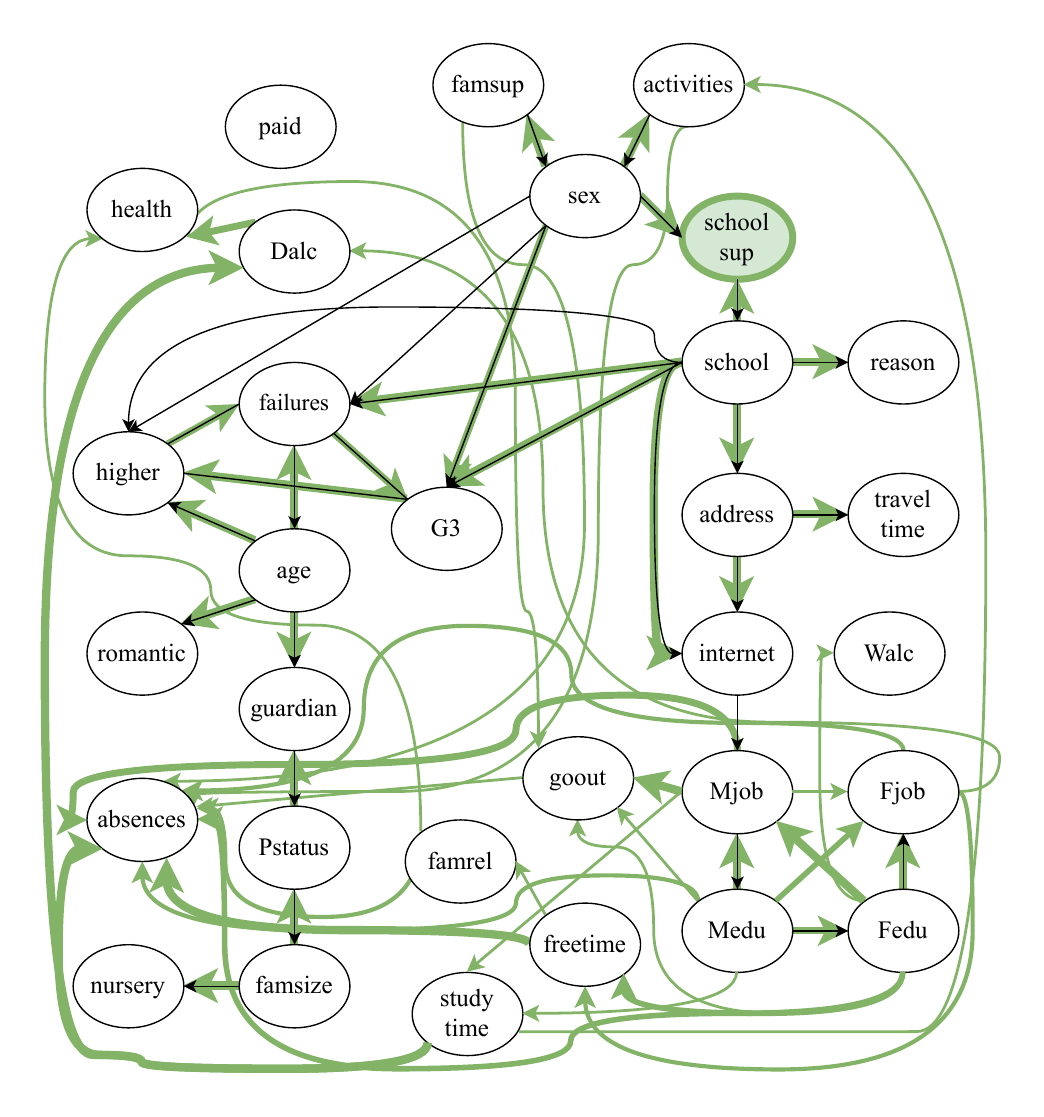}
        \caption{Student -- Girls Support}\label{fig:res-girls}
    \end{subfigure}
    \begin{subfigure}[b]{0.5\textwidth}
        \includegraphics[width=\textwidth]{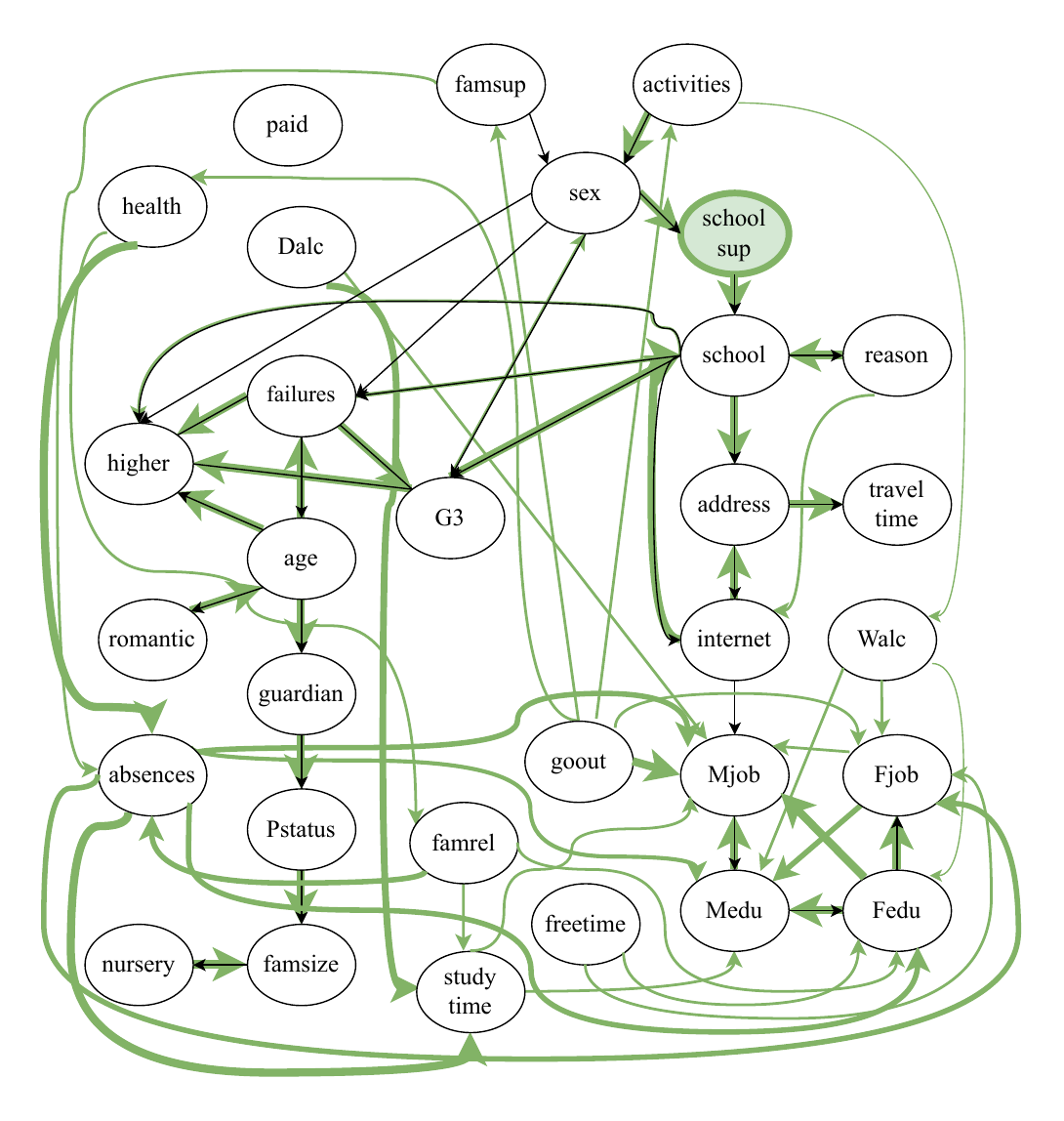}
        \caption{Student -- Boys Support}\label{fig:res-boys}
    \end{subfigure}
    \caption{Case Studies on the Student dataset. Black edges indicate the ground truth, while green edges indicate detections by the PC algorithm; thickness correlates with the number of runs in which an edge was detected. Children of $T$ are marked green, with the thickness of the border indicating the number of runs that correctly identified this relationship.}
\end{figure}

Finally, despite the poor overall performance of the PC algorithm on the \emph{Student Performance} dataset, our drift scenario \emph{Student Girls Support} (\cref{fig:res-girls}) shows that our causal explanation framework works reliably even when large parts of the causal graph are poorly recovered. In all ten runs, the drifting feature \emph{schoolsup}, short for "school support", is correctly identified as the only child of $T$, and the conditional variable \emph{sex} is identified as the only other parent of the drifting feature. This shows that even under less-than-ideal circumstances, which cause the PC algorithm to largely fail overall, causal explanations of concept drift are reliably extracted by our method.
Similar findings can be seen in the \emph{Student Boys Support} scenario (\cref{fig:res-boys}), where an analogous causal structure may be observed. This implies that in scenarios where concept drift is sufficiently strong, causal explanations work well despite the high number of features and the low number of data points.

\section{Conclusion}
\label{sec:concl}
In this paper, we proposed a method for causally explaining concept drift that directly enables the user to perform drift-reversing interventions on the system at hand. Our methodology is integrated into the model-based drift explanation framework and, thus, leverages the modelling of drift as a dependence of data and time. By incorporating computational causality, we can identify the full set of drift-reversing interventions. Since this may be too extensive for a layperson, we introduced conditional interventions and, thereby, obtained actionable explanations. We experimentally showed that even though the PC algorithm does not always yield a reliable causal graph of the data at hand, our pipeline reliably identifies features directly impacted by the drift.

As mentioned above, related work on incorporating feature relevance theory into model-based drift explanations seems to yield similar explanations. Investigating the relationship between causal discovery algorithms and feature relevance in this context seems to be an interesting further step. In particular, to reduce the amount of data needed. 
So far, our considerations assume that the entire dataset can be described by one causal graph. However, in some real-world settings, the assumption that one causal graph describes the data globally does not hold. In these cases, finding subgroups in the data and providing more local explanations for specific populations would be advantageous.

\begin{credits}
\subsubsection{\ackname} Funding in the scope of the BMBF project KI Akademie OWL under grant agreement No 01IS24057A and the ERC Synergy Grant ``Water-Futures'' No. 951424 is gratefully acknowledged.

\subsubsection{\discintname} The authors have no competing interests to declare that are relevant to the content of this article.
\end{credits}